\documentclass[10pt,twocolumn,letterpaper]{article}

\usepackage[pagenumbers]{wacv} 

\usepackage{graphicx}
\usepackage{amsmath}
\usepackage{amssymb}
\usepackage{booktabs}

\usepackage{adjustbox}
\usepackage{xcolor}
\usepackage{booktabs}
\usepackage{subcaption}
\usepackage{makecell}
\usepackage{multirow}
\usepackage{tabularx}

\usepackage[pagebackref,breaklinks,colorlinks]{hyperref}

\usepackage[capitalize]{cleveref}
\crefname{section}{Sec.}{Secs.}
\Crefname{section}{Section}{Sections}
\Crefname{table}{Table}{Tables}
\crefname{table}{Tab.}{Tabs.}

\def\wacvPaperID{1272} 
\def\confName{WACV}
\def\confYear{2025}

\begin{document}

\title{Anomaly Detection for People with Visual Impairments \\Using an Egocentric 360-Degree Camera}

\author{
    Inpyo Song$^1$, Sanghyeon Lee$^2$, Minjun Joo$^1$, Jangwon Lee$^1$\\
    $^1$Department of Immersive Media Engineering, Sungkyunkwan University\\
    $^2$School of Electronics and Information Engineering, Korea Aerospace University\\
    \tt\small{\{songinpyo, jmjs1526, leejang\}@skku.edu, tkdgus4693@kau.kr}
}

\maketitle

\begin{abstract}
Recent advancements in computer vision have led to a renewed interest in developing assistive technologies for individuals with visual impairments.
Although extensive research has been conducted in the field of computer vision-based assistive technologies,
most of the focus has been on understanding contexts in images,
rather than addressing their physical safety and security concerns.
To address this challenge, we propose the first step towards detecting anomalous situations for visually impaired people by observing their entire surroundings using an egocentric 360-degree camera.
We first introduce a novel egocentric 360-degree video dataset called
\textbf{VIEW360} (Visually Impaired Equipped with Wearable 360-degree camera),
which contains abnormal activities that visually impaired individuals may encounter, such as shoulder surfing and pickpocketing.
Furthermore, we propose a new architecture called the \textbf{FDPN} (Frame and Direction Prediction Network),
which facilitates frame-level prediction of abnormal events and identifying of their directions.
Finally, we evaluate our approach on our VIEW360 dataset and the publicly available UCF-Crime and Shanghaitech datasets,
demonstrating state-of-the-art performance.
\end{abstract}

\section{Introduction}
\label{sec:intro}

People with visual impairments encounter various challenges in their daily lives, especially related to their physical safety and security risks,
as they may not perceive their surroundings as easily as sighted individuals \cite{ahmed2016addressing}.
Traditionally, white canes and guide dogs have been widely used to help them navigate their environment and understand their surroundings.
However, these traditional assistive systems suffer from certain limitations.
For example, white canes only provide limited information about the surroundings \cite{dos2021electronic},
and training guide dogs is both time-consuming and expensive \cite{pfaffenberger1976guide}.
As a result, there has been a significant increase in developing a visual aid system to create new ``eyes''
for visually impaired individuals using wearable devices and Artificial Intelligence (AI) technologies \cite{sivan2016computer, gurari2018vizwiz}.
However, much of the research up to now has primarily focused on tasks
like the image captioning and visual question answering
rather than addressing real-world issues such as physical safety and security concerns
\cite{gurari2018vizwiz, gurari2019vizwiz, gurari2020captioning}.
Therefore, this paper proposes a first step towards
to address the physical safety and security concerns of visually impaired people
by employing a 360-degree wearable camera.
In particular, we have two primary objectives:
(1) detecting suspicious or abnormal activities within a 360-degree video stream,
(2) identifying the direction of these activities.

\begin{figure}[!t]
\includegraphics[width=\columnwidth]{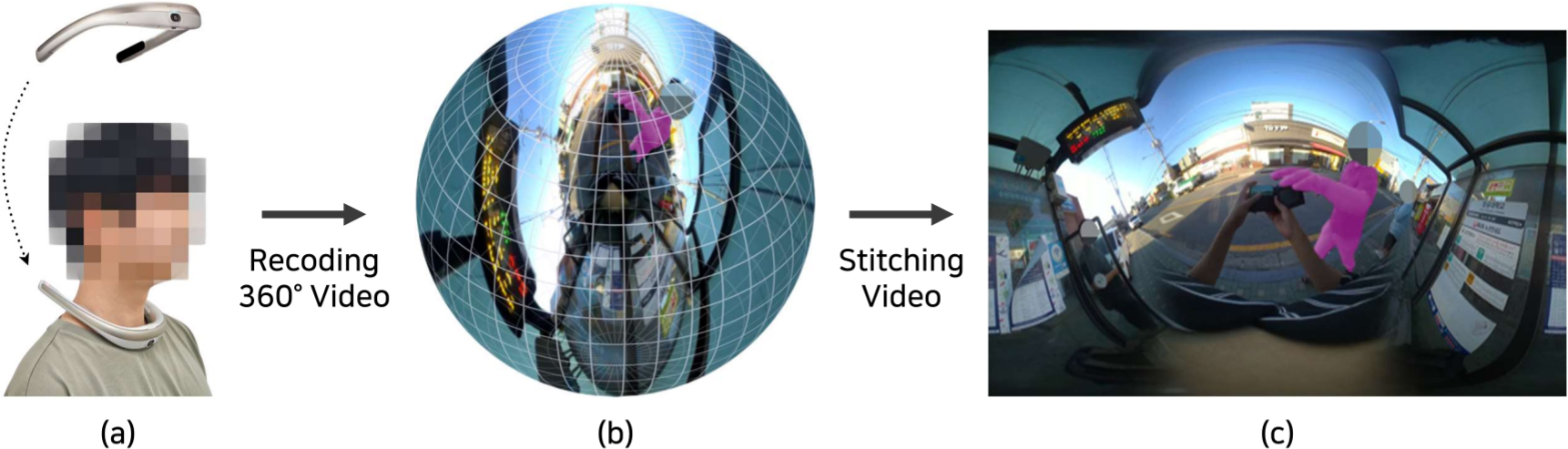}
\caption{
This paper aims to tackle safety and security concerns faced by visually impaired individuals.
To tackle these concerns, we introduce a new dataset, \textbf{VIEW360},
specifically designed for detecting unusual activities
by observing their entire surroundings using an egocentric 360-degree camera.
The dataset is collected through a process involving 
(a) capturing footage with a wearable 360-degree camera worn around the neck,
(b) recording egocentric 360-degree videos to encompass the wearer's surroundings,
and (c) stitching these videos into panoramic views for comprehensive analysis.
In the depicted scene, the individual highlighted in magenta is attempting a wallet theft.
}
\vspace{-1em}
\label{fig:front_image}
\end{figure}

To accomplish this, we first introduce a new dataset called
\textbf{VIEW360},
which has been designed for detecting anomalies in a camera wearer's entire surroundings.
This dataset includes egocentric 360-degree videos captured at several public locations,
such as ATM booths, parks, and cafes.
VIEW360 is comprised of 575 videos that illustrate real-life situations,
which were obtained through interviews with visually impaired individuals \cite{ahmed2016addressing}.
Our dataset can be categorized as one of the datasets for the Video Anomaly Detection (VAD) task in the field,
which aims to identify frames in a video where unusual or anomalous events occur.
There are generally two approaches to this task: semi-supervised Video Anomaly Detection (\textbf{sVAD}) and weakly-supervised Video Anomaly Detection (\textbf{wVAD}).
The sVAD approach trains networks using only normal videos without any annotations
\cite{sabokrou2018adversarially, wang2022video},
whereas the wVAD approach uses video-level labels for training, but with limited annotation
\cite{sultani2018real, tian2021weakly}.

Among the two lines of approaches, lately, 
researchers have shown an increased interest in wVAD methods
due to their promising performance on public VAD benchmark datasets \cite{feng2021mist, chen2023mgfn}.
However, these methods face challenges in identifying abnormal activities in rapidly changing scenarios.
This limitation is inherent to recent wVAD approaches,
which tend to predict anomaly scores at the snippet-level (a brief segment extracted from a video).
Consequently, these approaches assign identical anomaly scores to a fixed number of frames
(referred to as snippets), often resulting in overly generalized predictions.
Therefore, these techniques could face difficulties in identifying sudden real-world anomalies,
such as short and unexpected activities like shoulder surfing at an ATM booth.

To address these limitations, we introduce a framework called Frame and Direction Prediction Network (\textbf{FDPN}).
It is designed to predict anomaly scores at the frame level, extending beyond mere snippet-level scores.
We achieve this through our innovative coarse-to-fine learning approach.
By utilizing existing snippet-level predictions \cite{chen2023mgfn, tian2021weakly} as pseudo-supervision,
our FDPN is trained to predict frame-level anomalies, 
eliminating the requirement for extra annotations.
This frame-level prediction proves particularly advantageous in identifying abrupt abnormal events
occurring within a brief timeframe, as depicted in Figure~\ref{fig:frame_video_comparision}.

Furthermore, we employ an off-the-shelf saliency detection model as our pre-processing step,
denoted as saliency-driven image masking, to further enhance the process.
This technique identifies visually striking regions in a frame that may contain anomalies.
Since anomalies often appear in salient regions, this method allows us to narrow down the search space for anomaly detection.
This is especially useful for handling 360-degree images due to their extensive visual coverage.
We also incorporate direction classification from a 360-degree egocentric perspective, offering practical guidance for visually impaired individuals during anomalous events.
This is achieved through a dedicated subnetwork utilizing saliency heatmaps.

Finally, we evaluate the proposed approach on our VIEW360 and the publicly available UCF-Crime, Shanghaitech datasets.
Our approach achieves state-of-the-art performance on the VIEW360, UCF-Crime and Shanghaitech datasets. In summary, this paper makes the following contributions:
\begin{itemize}
  \setlength\itemsep{0em}
  \item To our knowledge, this is the first study to address the physical safety and security concerns of people with visually impairments by detecting anomalous events and identifying their direction in the surroundings.
  \item We introduce \textbf{VIEW360}, a novel egocentric 360-degree video-based dataset to address safety and security concerns of visually impaired people in real-world scenarios.
  \item We propose a novel architecture called \textbf{FDPN} that can predict more precise anomaly scores at the frame-level based on rich scene representation without the need for additional frame-level annotation.
\end{itemize}

\section{Related Work}
\subsection{AI for People with Visual Impairments}
In the last decade, there has been a dramatic increase in developing a visual aid system aimed at creating
new ``eyes'' for the visually impaired people using AI technologies.
Gurari et al. have introduced the VizWiz-VQA dataset,
a collection of images and questions gathered from blind individuals \cite{gurari2018vizwiz}.
These researchers have also created another VQA dataset, named the VizWiz-Priv dataset,
which focuses on identifying unintended leaks of personal information through VQA for visually impaired users \cite{gurari2019vizwiz}.
Furthermore, diverse approaches have emerged in recent years to offer various perspectives within this VQA task.
These include delving into the reasons behind variations in responses to identical visual questions
among distinct individuals \cite{bhattacharya2019does},
addressing the domain gap between images captured by visually impaired individuals
and sighted individuals  \cite{gurari2020captioning},
as well as tackling poor image quality in VQA systems
by creating a dataset and task to predict reasons for low-quality images \cite{chiu2020assessing}.

While existing efforts have made substantial progress in comprehending the contextual aspects of images,
less attention has been directed towards video analysis,
especially in addressing the security and physical concerns that visually impaired individuals might encounter in their daily lives.
Therefore, this study bridges a research gap in AI-based visual aid systems for people with visual impairments
by introducing a novel 360-degree egocentric video anomaly detection task along with a new dataset.

\subsection{Anomaly Detection in Videos}
Video anomaly detection, a crucial computer vision task, identifies frames with abnormal events in videos.
Approaches are categorized as semi-supervised and weakly-supervised.
Semi-supervised methods, assuming most data is normal, detect abnormalities
by identifying image frames that significantly differ from previously observed data
\cite{sabokrou2018adversarially, nguyen2019anomaly}.
This approach makes sense since abnormal events ``unusually'' occur in the real world
and it offers a clear advantage as it does not require annotation costs.
However, their performance often declines when classifying unseen data.

In contrast, weakly-supervised approaches aim to enhance detection performance by utilizing minimal annotations,
such as video-level labels, during training.
Recent advancements in this field include the use of graph networks to handle noisy labels in abnormal videos \cite{zhong2019graph}, and the integration of motion information to improve detection accuracy \cite{zhu2019motion}.
Further innovations, such as inter-class distancing, sequence learning, self-supervised techniques, and magnitude-based methods, have significantly contributed to the progress in anomaly detection \cite{wan2020weakly, li2022self, wu2022self, tian2021weakly, chen2023mgfn}. Additionally, the adoption of vision-language models and prompt-enhanced techniques has advanced capabilities in this domain \cite{joo2023clip, yang2024text, wu2024vadclip, chen2024prompt}.
However, existing methods, which are primarily snippet-level anomaly detection approaches, often struggle with abrupt events due to uniform scoring. Our proposed frame-level approach leverages snippet-level predictions as pseudo-supervision, thereby improving detection accuracy for short-lived anomalies.

\begin{figure}[t]
\centering
\includegraphics[width=\columnwidth]{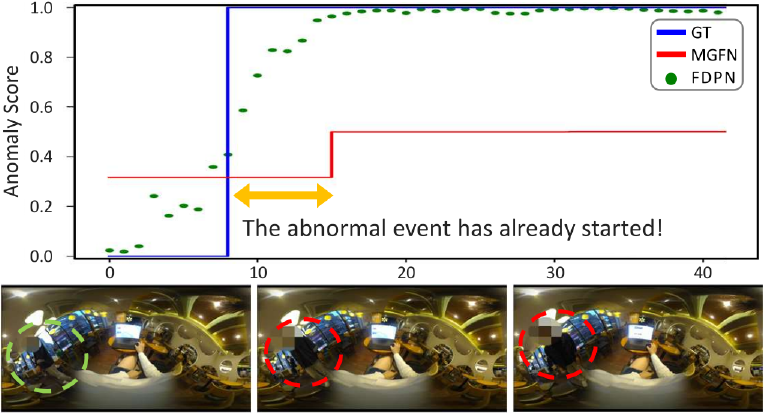}
\caption {
This figure contrasts anomaly scores at event start and end boundaries
for state-of-the-art method MGFN and our FDPN on VIEW360 dataset.
MGFN often makes false predictions at event boundaries because it predicts at the snippet-level,
whereas our proposed method makes better predictions at the event boundaries
since it can make frame-level predictions.
}
\vspace*{-1em}
\label{fig:frame_video_comparision}
\end{figure}

\begin{figure*}[!t]
\centering
\includegraphics[width=\textwidth, height=14em]{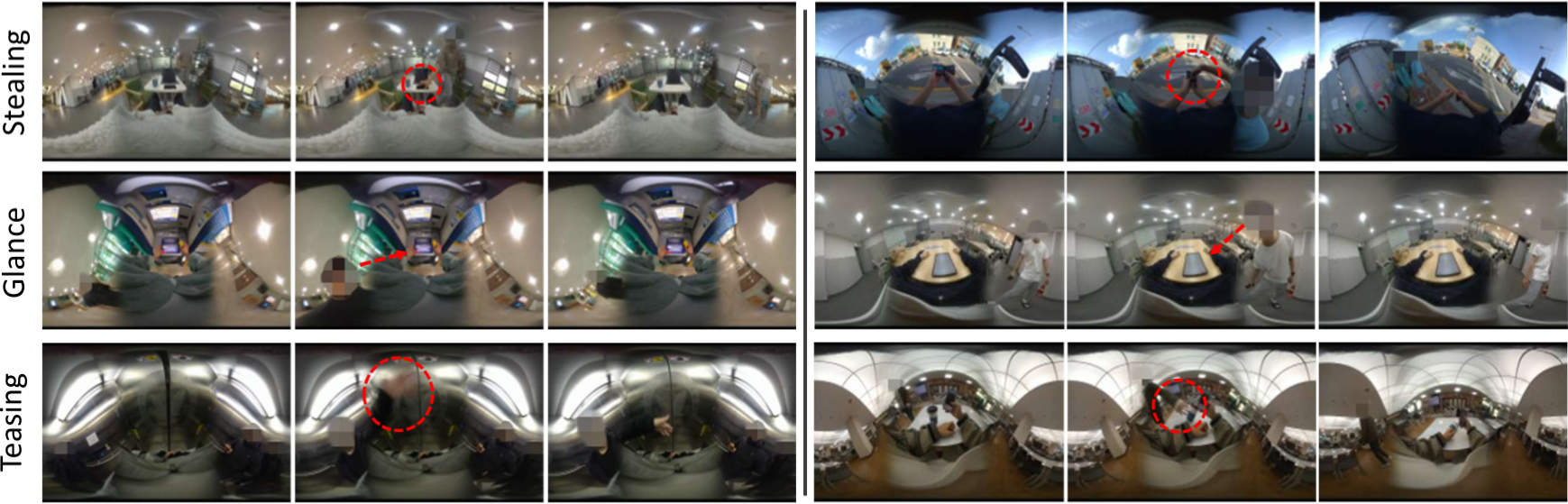}
   \caption{
Here are some abnormal instances in our VIEW360 dataset.
The first row shows theft of personal belongings from the camera-wearer.
The second row depicts shoulder-surfing attacks:
someone covertly observing the camera wearer's ATM use and smartphone without their awareness.
The third row portrays a person with visual impairments being mocked or harassed.
}
\label{fig:dataset-images}
\vspace*{-1em}
\end{figure*}

\section{VIEW360:Dataset for the Visually Impaired} \label{sec:view360}
This section presents the ``\textbf{V}isually \textbf{I}mpaired \textbf{E}quipping \textbf{W}earable \textbf{360}-degree camera'' (\textbf{VIEW360}) dataset,
designed to advance AI-assisted technology for individuals with visual impairments
through a 360-degree wearable camera \cite{fitt360}.
This is the first dataset that contains 360-degree egocentric videos of anomaly activities
that visually impaired people commonly encounter in their daily lives.
We selected the locations and anomalies for our dataset
through interviews with visually impaired individuals \cite{ahmed2016addressing}.
We focused on three types of abnormal scenarios: \textit{Glance, Stealing}, and \textit{Teasing}, as illustrated in Figure~\ref{fig:dataset-images}.
Data was collected from eight different public venues: \textit{Cafes, Restaurants, Bus stops, Elevators, Parks, Libraries, Offices}, and \textit{Automated Teller Machine (ATM) booths}.

\noindent \textbf{Video Collection}
All videos in the dataset were captured using a 360-degree wearable camera mounted on
the neck of an actor who simulated being visually impaired.
To ensure diversity, 11 participants wore the camera and performed abnormal behaviors.
We aimed to construct a dataset with a range of situations,
so we mostly collected short videos ranging from 10 to 60 seconds
instead of long videos typical of most video anomaly datasets.
In total, we collected 575 videos consisting of 484,364 frames.

\noindent \textbf{Annotations}
There are two types of annotations available: temporal annotations and directional annotations.
Temporal annotations are created in accordance with the conventions of existing anomaly detection datasets \cite{sultani2018real, wu2020not}.
Specifically, we annotated video-level labels for the training set, while the testing set includes frame-level labels for evaluation purposes.
A challenge in labeling the anomaly detection dataset is determining the boundary between the beginning and end of the anomaly.
To address this, a total of five annotators were engaged, cross-validating each other's labels to maintain consistency and enhance accuracy.
For direction prediction, we deliberately simplified the annotation to three categories, \textit{Left back, Center}, and \textit{Right back}.
This sparse but practical directional information can aid visually impaired individuals in swiftly identifying potential threat directions,
enabling quicker decision-making and more effective responses in real-world scenarios.

{
\setlength\tabcolsep{6pt}
\begin{table}[t]
\centering
\small
\begin{adjustbox}{max width=\columnwidth}
\begin{tabular}{ l | c | c | c }
\hline
Dataset         & Videos & Avg. anomaly duration (s) & Video source \\
\hline
UCSD Ped1 \cite{li2013anomaly}      & 70     & 11.2 & CCTV \\
UCSD Ped2 \cite{li2013anomaly}       & 28     & 13.7  & CCTV \\
Avenue \cite{lu2013abnormal}     & 37     & 9.2 & CCTV \\
UBnormal \cite{acsintoae2022ubnormal} & 543 & 10.7 & Virtual scene \\
Shanghaitech \cite{liu2018future}    & 437    & 6.7 & CCTV \\
NWPU Campus \cite{cao2023new} & 547 & 10.8 & CCTV \\
\hline
\hline
UCF-Crime \cite{sultani2018real}       & 1,900  & 20.1 & CCTV \\
XD-violence \cite{wu2020not}     & 4,754  & 37.5 & Movie, Game, etc.\\
\textbf{VIEW360 (Ours)}   & 575    & 3.5 & Ego 360° camera\\
\end{tabular}
\end{adjustbox}
{
\caption{
Comparison of anomaly detection datasets, distinguishing between sVAD (top) and wVAD (bottom).
Our VIEW360 dataset, stands out by exclusively featuring egocentric 360-degree videos.
Notably, VIEW360 focuses on shorter average anomaly duration, emphasizing the detection of quick, transient anomalies.
}
\vspace{-1em}
\label{tab:VAD dataset comparision}
}
\end{table}
}

\begin{figure}[!t]
\centering
\includegraphics[width=0.9\columnwidth]{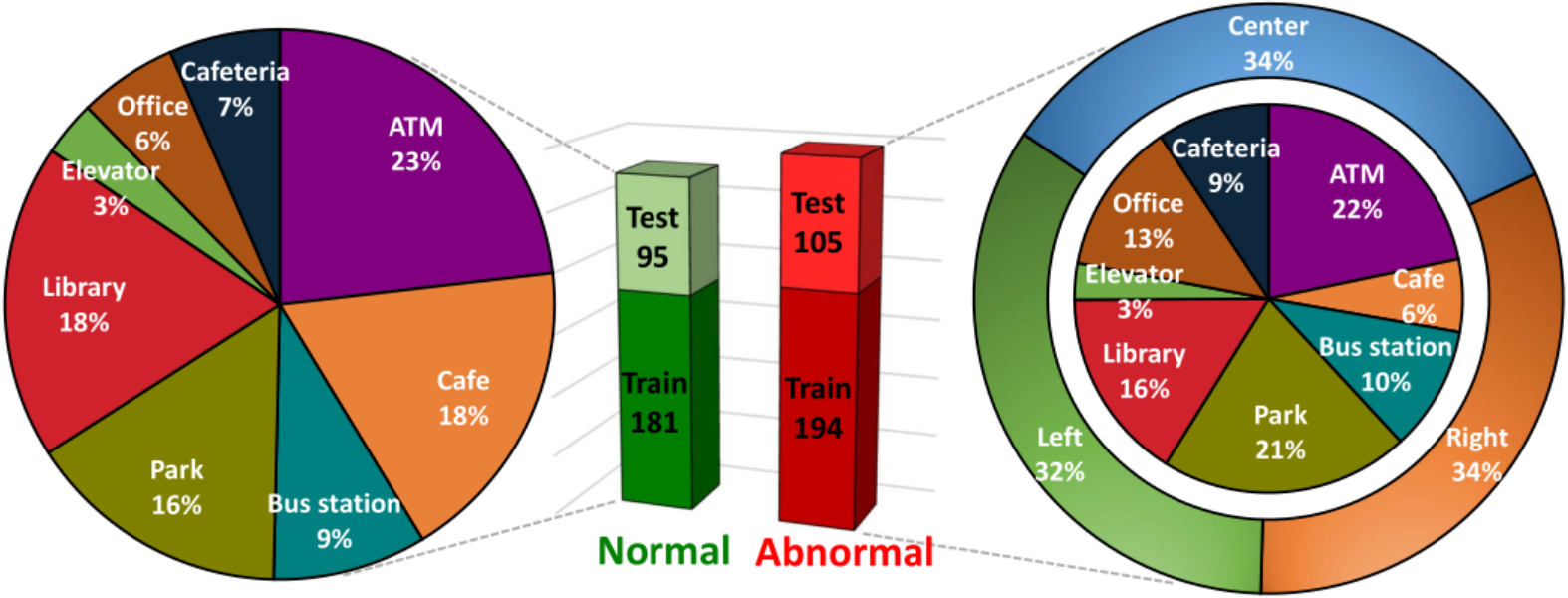}
   \caption{
   Distribution of the VIEW360 dataset, illustrating training/testing splits, video locations, and abnormal event orientations. Includes a bar chart of normal and abnormal video counts, a pie chart of video location distribution, and a donut chart of abnormal event directions.
   }
\vspace{-1em}
\label{fig:Dataset_distribution}
\end{figure}

\begin{figure}[!t]
\centering
\includegraphics[width=\columnwidth]{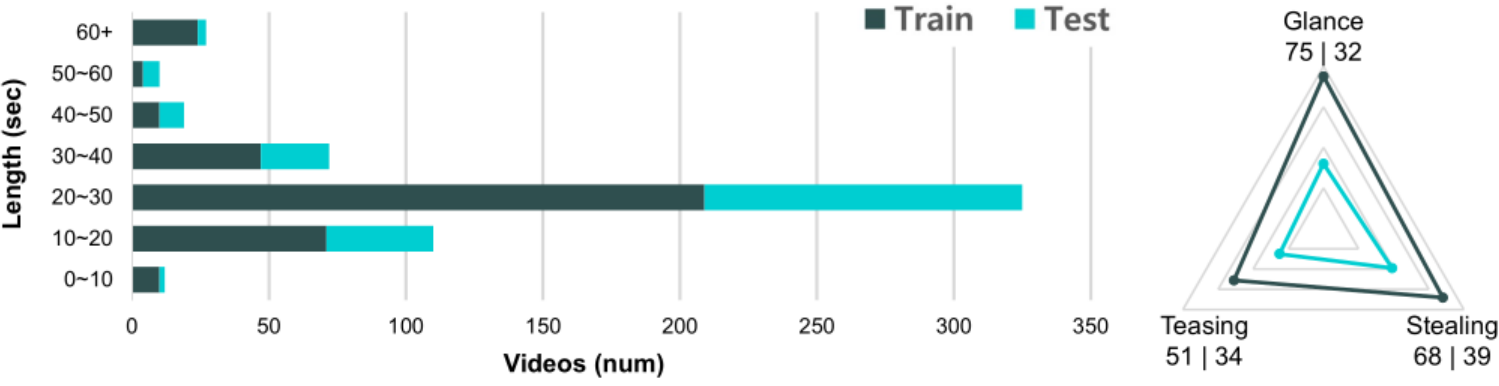}
   \caption{
   Video duration and abnormal classes in VIEW360.
   }
\vspace{-1em}
\label{fig:data_length}
\end{figure}

\noindent \textbf{Dataset Statistics}
The training set comprises 375 videos: 181 normal and 194 abnormal.
The testing set includes 200 videos: 95 normal and 105 abnormal.
Our dataset features 3 directional labels,
\textit{Left back} (106 videos), \textit{Center} (101 videos), and \textit{Right back} (105 videos).
Details of training/testing sets, locations, and abnormal directions are in Figure~\ref{fig:Dataset_distribution}.
Abnormal situations are evenly distributed between the training and testing sets.
Video length (in seconds) and abnormal class distribution are in Figure~\ref{fig:data_length}.

\noindent \textbf{Privacy and Ethics}
The dataset collection for the VIEW360 was rigorously conducted under the approval of the Institutional Review Board (IRB), ensuring that all research activities adhered to the highest ethical standards and guidelines.
Informed consent was obtained from all participants involved in the simulated scenarios, thereby guaranteeing their full awareness of the data collection's purpose and scope, and affirming their rights as participants.
During the collection process, special attention was given to respecting the rights of others in private spaces and diligently avoiding the capture of sensitive areas or activities.
For non-consenting individuals appearing in public areas within the videos, privacy measures like blurring and facial masking were applied to uphold their anonymity and ensure the dataset adhered to ethical research practices.

\section{Proposed Approach}
Building upon our VIEW360 dataset,
our goal is to identify short-lived abnormal activities in a 360-degree video stream
and determine their direction.
In this section, we introduce our framework, the Frame and Direction Prediction Network (FDPN), designed to achieve these aims.

\begin{figure*}[t]
\centering
\includegraphics[width=\textwidth]{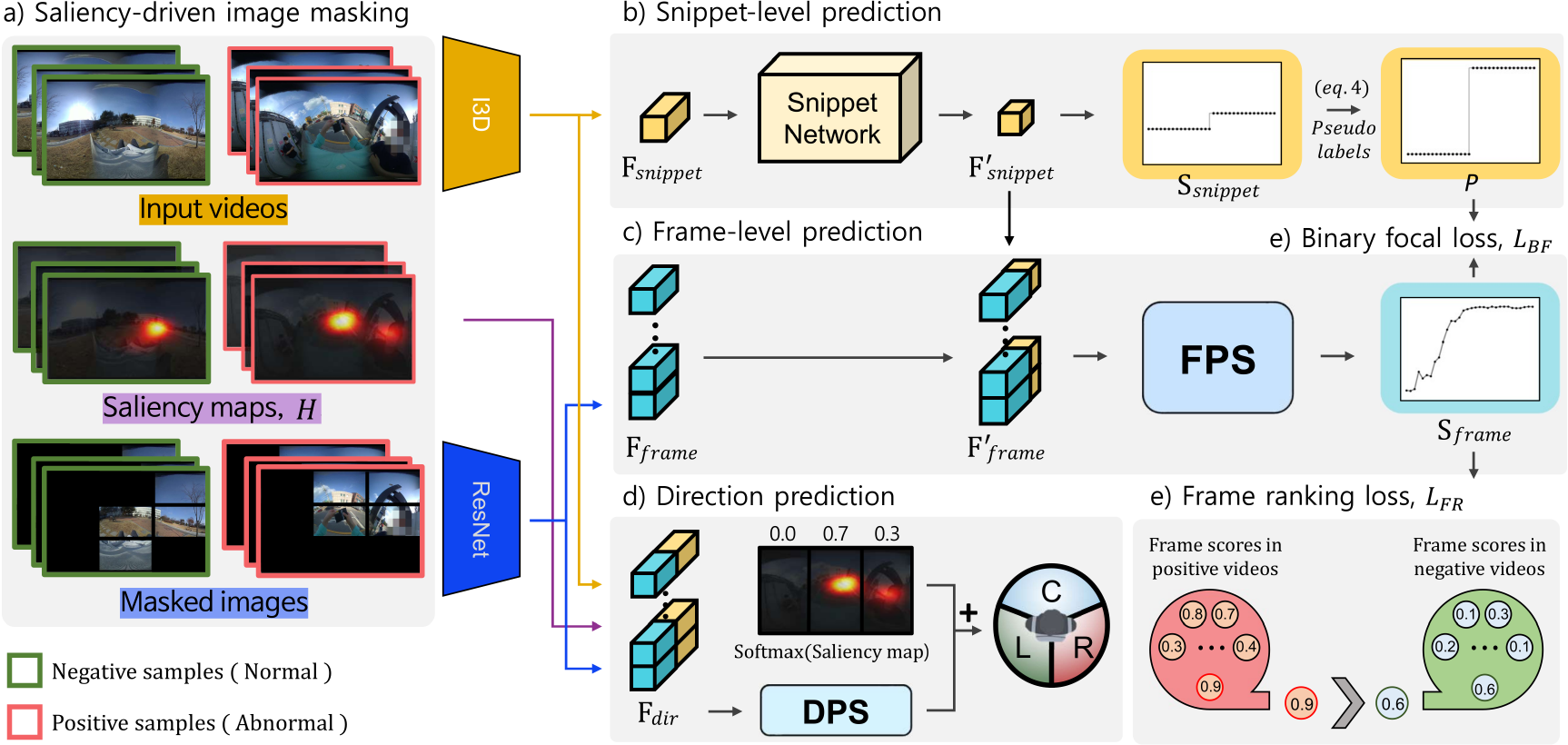}
\caption{
Overview of our FDPN.
During training, positive and negative video pairs are fed into the framework.
\textbf{(a)} Saliency map-based masking is applied to these pairs,
identifying salient regions within input image frames.
Snippet-level original frames are processed using an I3D feature extractor,
while frame-level masked images are handled by ResNet.
\textbf{(b)} Following feature extraction, the Snippet Network generates $F'_{\text{snippet}}$
and computes snippet-level anomaly score $S_{\text{snippet}}$,
used to create a pseudo label ($\mathcal{P}$) for training the FPS.
\textbf{(c)} FPS employs the concatenated feature $F'_{\text{frame}}$,
combining $F_{\text{frame}}$ from masked images and $F'_{\text{snippet}}$,
to compute frame-level anomaly score $S_{\text{frame}}$.
\textbf{(d)} For direction estimation, we construct the DPS.
DPS initially estimates direction using concatenated feature $F_{\text{dir}}$,
refining output with softmax-applied saliency values.
\textbf{(e)}
To train the model, we use binary focal loss with $S_{\text{frame}}$ and the pseudo labels ($\mathcal{P}$).
In addition, Frame Ranking Loss ($L_{FR}$)
ensures higher anomaly scores for positive videos than for negative ones.
}
\label{fig:overall_architecture}
\vspace*{-1em}
\end{figure*}

\subsection{Overview}
Our FDPN begins by identifying salient regions within the input frame that may contain anomalies
given 360-degree panoramic image frames as input.
Once the salient regions in the image frames are identified, FDPN internally employs two types of input images:
1) masked images, retaining only the salient regions while masking out other portions, and
2) original images, the unaltered 360-degree panorama input images.
FDPN then proceeds to extract snippet-level features from the original images using Inflated 3D ConvNet (I3D) \cite{carreira2017quo},
and it extracts frame-level (image-level) features from the masked images using ResNet \cite{he2016deep}.
After that, we employ the snippet-level features for prediction at the snippet level and
to generate pseudo-labels for training our Frame Prediction Subnetwork (FPS).
Finally, these snippet-level features are combined with the frame-level features,
and this concatenated feature set is utilized to compute anomaly scores at the frame level.

For identifying the direction of abnormal events, we construct the Direction Prediction Subnetwork (DPS).
It operates on concatenated features of snippet-level and image-level features,
along with the saliency maps used in the initial step.
Figure~\ref{fig:overall_architecture} depicts the overall architecture of the FDPN.
We present the details of each module within this process in the following subsections.

\subsection{Saliency-driven Image Masking (a)}
\begin{figure}[t]
\centering
\includegraphics[width=\columnwidth]{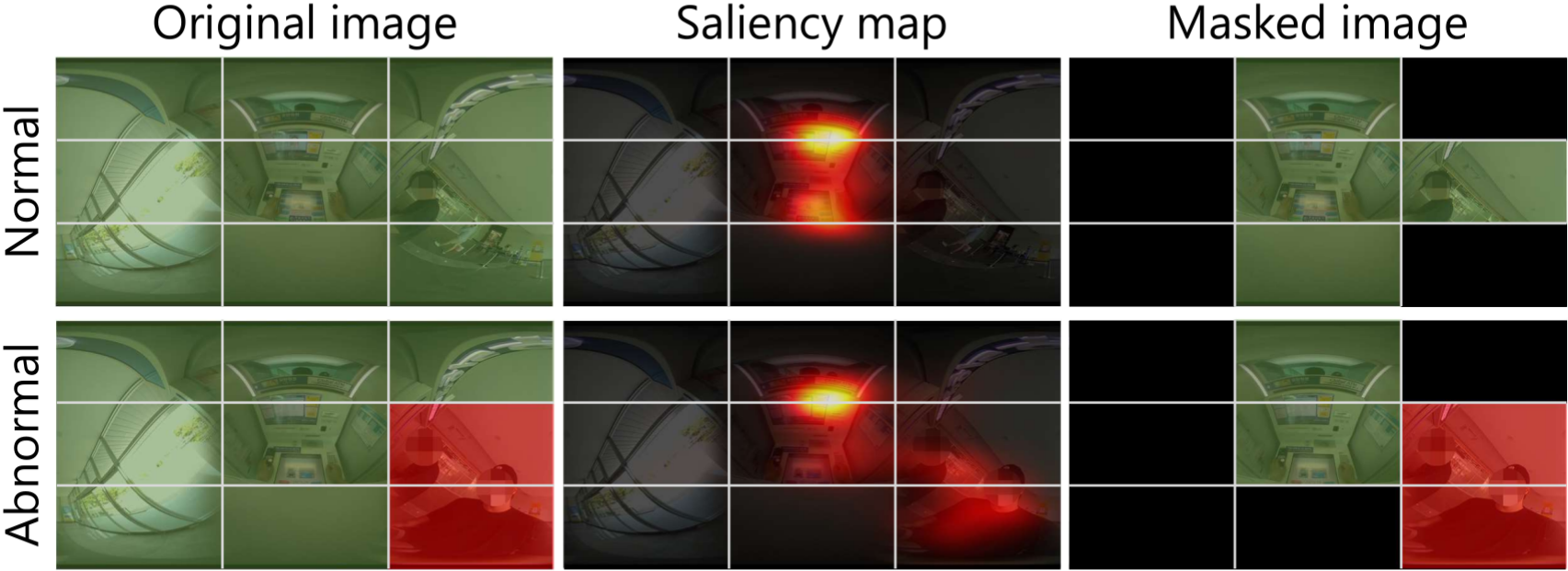}
\caption{
This figure shows that our saliency-driven image masking preprocess applied to events at an ATM booth.
It is divided into two rows, representing normal and abnormal events, with each row consisting of the original image, saliency map, and masked image, respectively.
In the bottom row representing abnormal events, we marked the grid cells where the anomaly activity happened in red.
   }
\label{fig:maksed-image}
\vspace{-1em}
\end{figure}

We first employ TASED-Net \cite{min2019tased} to derive saliency heatmaps $H$ as shown in Figure~\ref{fig:overall_architecture}-(a).
These heatmaps emphasize visual significance within the frame, with each pixel value $H_{x,y}$ representing the saliency score at the corresponding coordinates (x,y) within the image. 
Subsequently, we divide these heatmaps into an $n \times n$ grid.
For each cell denoted by integers $(i,j)$,
we then compute the importance score by summing the pixel saliency values within it:
\begin{equation}
G_{i,j} = \sum_{x,y \in \text{cell}_{i,j}} H_{x,y}
\end{equation}
After this step, we pinpoint the \textit{top-K} salient regions with the highest scores from the grid scores $G_{i,j}$.
Following this, we generate a binary mask $M$ where the \textit{top-K} cells are assigned a value of 1,
and the remainder are assigned a value of 0:

\begin{equation}
{M}_{i,j} =
\begin{cases} 
      1 & \text{if } G_{i,j} \text{ is in} \textit{ top-K} \\
      0 & \text{otherwise}
\end{cases}
\end{equation}
Finally, the masked image $I_{\text{masked}}$ is obtained by multiplying the original image $I$ by the binary mask $M$:
\begin{equation}
I_{\text{masked}} = I \odot M
\end{equation}
This approach offers a significant advantage:
it guides focus toward event-specific regions by utilizing the saliency map to pinpoint critical areas while concealing others.
This emphasis improves the analysis of potential anomaly locations, as shown in Figure~\ref{fig:maksed-image}.

From $I_{\text{masked}}$, frame-level features are extracted using ResNet,
yielding $F_{\text{frame}} \in \mathbb{R}^{B \times T \times N \times C}$.
Simultaneously, snippet-level features are derived from the original frames using I3D, resulting in $F_{\text{snippet}} \in \mathbb{R}^{B \times T \times C}$.
Here, $B$ represents the number of video pairs, $T$ the number of snippets, each snippet consists of $N$ frames and $C$ the channel.

\subsection{Coarse-to-Fine Learning (b - c)}

\noindent \textbf{Snippet-level Prediction (b)}
To enable our FDPN to produce frame-level anomaly scores,
the network needs frame-level ground truth labels for training.
Consequently, we initially train the Snippet Network for snippet-level prediction.
This involves using pre-trained networks \cite{tian2021weakly, chen2023mgfn} 
to create feature $F'_{\text{snippet}}$
and snippet-level anomaly scores $S_{\text{snippet}}$.
Subsequently, the snippet-level anomaly score $S_{\text{snippet}}$ is duplicated $N$ times,
following the specified rule below, resulting in pseudo labels $\mathcal{P}$ crafted for training our FPS at the frame level.
\begin{equation}
\mathcal{P} = 
\begin{cases}
0 & \text{if } S^{+}_{\text{snippet}} < 0.5 \text{ or } S^{-}_{\text{snippet}} \\
1 & \text{if } S^{+}_{\text{snippet}} \geq 0.5
\end{cases},
\label{eq1:Pseudo-label}
\end{equation}
$S^{+}_{\text{snippet}}$ and $S^{-}_{\text{snippet}}$ denote snippet-level anomaly scores
for positive and negative videos, respectively.
Positive videos receive pseudo labels of 0 or 1 according to their anomaly scores,
while negative videos (without anomalies) are assigned a pseudo label of 0.

\noindent \textbf{Frame-level Prediction (c)}
With the acquired pseudo labels, we proceed to train the FPS.
By combining $F_{\text{frame}}$ and $F'_{\text{snippet}}$,
we create $F'_{\text{frame}}$ as FPS input.
This design aims to encompass both frame-level and snippet-level information.
Following training, FPS is primed to estimate frame-level anomaly scores,
denoted as $S_{\text{frame}} \in \mathbb{R}^{B \times T \times N}$.
We designed this FPS to handle sequences, inspired by PoolFormer \cite{yu2022metaformer}.
It employs average pooling and 1D convolutions to capture frame connections and compute frame-level anomaly scores.
This improves the network's capability to understand relationships among neighboring frames.

\subsection{Direction Prediction (d)}
We also developed the DPS
to determine abnormal event directions by utilizing $F_{\text{dir}}$,
a concatenation of $F_{\text{frame}}$ and $F_{\text{snippet}}$.
DPS employs the same PoolFormer-based architecture as FPS and
computes direction scores for three orientations (Center, Left back, and Right back). 
It integrates saliency heatmap values across a 1x3 grid and applies softmax to enhance anomaly direction prediction.
This approach leverages spatial information from the saliency heatmap and our PoolFormer-based architecture
to improve the accuracy of abnormal event direction prediction, reducing spatial ambiguity in video frames.

\subsection{Loss Functions (e)}
We utilize three loss functions for frame-level prediction and a single loss function for direction prediction.
The first loss, Binary Focal Loss \cite{lin2017focal}, $\textit{L}_{BF}$,
compares the pseudo labels $\mathcal{P}$ to the frame-level anomaly scores $S_{\text{frame}}$
with a focusing parameter $\gamma$ that emphasizes hard-to-classify examples.

{\small
\begin{flalign}
\textit{L}_{BF} = -\mathcal{P}(1 - S_{\text{frame}})^\gamma \log(S_{\text{frame}}) \nonumber \\
-(1-\mathcal{P})S_{\text{frame}}^{\gamma} \log(1-S_{\text{frame}})
\label{eq2:Binary-Focal-Loss}
\end{flalign}
}
The second loss function, Frame Ranking Loss $\textit{L}_{FR}$,
prioritizes the top $\mathcal{R}$ frames with the highest anomaly scores in both positive and negative videos.
It drives the anomaly scores of positive videos towards 1 and those of negative videos towards 0.

{\small
\begin{flalign}
\textit{L}_{FR} = \frac{1}{\mathcal{R}}\sum_{r=1}^{\mathcal{R}} \left( 1 - S^{+}_{\text{frame}, r} + S^{-}_{\text{frame}, r} \right)
\label{eq3:Frame-Ranking-Loss}
\end{flalign}
}
The third loss, Smoothness Loss, $\textit{L}_{smooth}$ \cite{sultani2018real}, penalizes rapid changes in the predicted anomaly scores to ensure smooth transitions between adjacent frames.
It is calculated for all $\mathcal{F}$ frames as follows:

{\small
\begin{flalign}
\textit{L}_{smooth}=\frac{1}{\mathcal{F}}\sum_{f=1}^{\mathcal{F}} (S^{f}_{\text{frame}} - S^{f-1}_{\text{frame}})^{2}
\label{eq4:smoothness-Loss}
\end{flalign}
}

For direction prediction, we utilize Directional Focal Loss $\textit{L}_{DF}$,
employing the identical focusing parameter $\gamma$.
{\small
\begin{flalign}
\textit{L}_{DF} = -\sum_{k=1}^{C} y_{k}(1 - p_{k})^{\gamma} \log(p_{k})
\label{eq5:Direction-Focal-Loss}
\end{flalign}
}
Here, $C$ represents the three directions (Center, Left back, Right back) in our VIEW360 dataset.
$y_{k}$ is the ground-truth label for direction and $p_k$ is the predicted probability for that direction.
The parameter $\gamma$ is shared with Binary Focal Loss and adjusts the class contribution to the loss.

The overall loss is computed as follows,
where $\lambda_{1}$, $\lambda_{2}$, and $\lambda_{3}$
represent weight factors for each respective loss function.
{\small
\begin{flalign}
\textit{L} = \textit{L}_{BF} + \lambda_{1}\textit{L}_{FR} + \lambda_{2}\textit{L}_{smooth} + \lambda_{3}\textit{L}_{DF}
\label{eq4:Total-Loss}
\end{flalign}
}

\section{Experiments}
\subsection{Datasets and Evaluation Metrics}
We evaluate our method on three anomaly detection datasets: VIEW360, UCF-Crime, and Shanghaitech \cite{sultani2018real,liu2018future}.
We used Area Under the Receiver Operating Characteristic (AUC-ROC) evaluation metric same with established works \cite{tian2021weakly,chen2023mgfn}.
For VIEW360, we additionally employed AUC-ROC and Area Under the Precision-Recall (AUC-PR) metrics to evaluate model,
addressing both false positives and false negatives.
This approach ensures a balanced assessment of the model's ability to accurately detect anomalies while minimizing unnecessary alerts for visually impaired users, enhancing the system's practical reliability.

\subsection{Implementation Details}
Following existing methods \cite{chen2023mgfn,  wu2022self}, each video is divided into 32 snippets with 16 frames in each snippet during the training stage.
For the hyperparameters, we set $B=16$, $T=32$, $N=16$, $n=3$, $K=4$, $\gamma=2$, $\mathcal{R}=48$, $\lambda_{1}=1$, $\lambda_{2}=1.6e^{-3}$, $\lambda_{3}=0.3$ .

{
\setlength\tabcolsep{6pt}
\begin{table}[!t]
\centering
\resizebox{\linewidth}{!}{
\small
\begin{tabular}{ l | c | c | c | c}
Method & Publication & Feature & AUC-ROC & AUC-PR \\
\hline
MIST \cite{feng2021mist} & CVPR"21 & I3D & 79.30 & 19.58 \\
RTFM \cite{tian2021weakly} & ICCV"21 & I3D & 83.92 & 24.94 \\
S3R \cite{wu2022self}  & ECCV"22 & I3D & \underline{83.96} & \underline{25.17} \\
MGFN \cite{chen2023mgfn} & AAAI"23 & I3D & 80.43 & 20.16 \\
DMU \cite{zhou2023dual} & AAAI"23 & I3D & 83.88 & 25.11 \\
CLIP-TSA \cite{joo2023clip} & ICIP"23 & CLIP & 80.03 & 17.41 \\
VadCLIP \cite{wu2024vadclip} & AAAI"24 & CLIP & 79.92 & 21.28 \\
\hline
\textbf{FDPN} (Ours) & - & I3D & \textbf{86.00} & \textbf{26.97} \\
\end{tabular}
}
\caption{
Comparison of frame-level AUC-ROC and AUC-PR performances on VIEW360 dataset.
}
\label{tab:VIEW360 AUC}
\end{table}
}

{
\setlength\tabcolsep{6pt}
\begin{table}[!t]
\centering
\resizebox{\linewidth}{!}{
\small
\begin{tabular}{ l | c | c | c | c }
Method & Publication & Feature & UCF-Crime & Shanghaitech \\
\hline
MIST \cite{feng2021mist} & CVPR"21 & I3D        &  82.30 & 94.83\\
RTFM \cite{tian2021weakly} & ICCV"21 & VidSwin  &  83.31 & 96.76\\
RTFM \cite{tian2021weakly} & ICCV"21 & I3D        &  84.03 & 97.21\\
MSL \cite{li2022self}  & AAAI"22 & I3D        &  85.30 & 95.45 \\
MSL \cite{li2022self}  & AAAI"22 & VidSwin  &  85.62 & 96.93 \\
S3R \cite{wu2022self}  & ECCV"22 & I3D        &  85.99 & 97.48\\
MGFN \cite{chen2023mgfn} & AAAI"23 & I3D        &  86.98 & -\\
DMU \cite{zhou2023dual} & AAAI"23 & I3D       & 86.97 & - \\
CLIP-TSA \cite{joo2023clip} & ICIP"23 & CLIP & 87.58 & 98.32 \\
PE-MIL \cite{chen2024prompt} & CVPR"24 & I3D        &  86.83 & \underline{98.35}\\
\textbf{*}TPWNG \cite{yang2024text} & CVPR"24 & CLIP        &  87.79 & -\\
\textbf{*}VadCLIP \cite{wu2024vadclip} & AAAI"24 & CLIP        &  \underline{88.02} & - \\
\hline
\textbf{FDPN} (Ours) & - & I3D & \textbf{88.03} & \textbf{98.51} \\
\end{tabular}
}
\caption{
Comparison of frame-level AUC-ROC performance on UCF-Crime and Shanghaitech datasets.
Models marked with ``\textbf{*}" use multi-modal text information, which was not available for evaluation on the Shanghaitech dataset.
}
\label{tab:ucf-crime and shanghaitech}
\vspace{-1em}
\end{table}
}

\subsection{Evaluation}
\noindent \textbf{VIEW360}
As shown in Table~\ref{tab:VIEW360 AUC},
our approach outperforms state-of-the-art methods in both AUC-ROC and AUC-PR on the VIEW360 dataset.
We achieved improvements of 2.08\% and 2.03\% in AUC-ROC and AUC-PR, respectively,
compared to RTFM (used as the snippet network).
These results highlight the efficacy of our method in detecting subtle anomalies and short-life anomalies in the dataset,
important for applications with visually impaired users.

\noindent \textbf{UCF-Crime and Shanghaitech}
Our FDPN achieved state-of-the-art performance on UCF-Crime and Shanghaitech as shown in Table~\ref{tab:ucf-crime and shanghaitech}.
These results highlight the robustness of our frame-level prediction method across different anomaly detection tasks.

\noindent \textbf{Analysis of Existing Methods}
We observed an intriguing performance discrepancy between recent state-of-the-art methods like MGFN and VadCLIP across different datasets.
While these methods excelled on UCF-Crime, they underperformed on VIEW360.
This variance primarily stems from the inherent differences in dataset characteristics.
VIEW360 contains more subtle and shorter-duration anomalies compared to UCF-Crime.
Therefore, MGFN, optimized for high-magnitude snippet training,
tends to struggle with accurate clip selection in datasets featuring subtler anomalies like VIEW360.
Similarly, VadCLIP's reliance on abnormal category classification for training proves less effective for VIEW360,
where abnormal events are more challenging to distinguish than in UCF-Crime.

\noindent \textbf{Snippet Network Selection}.
We tailored our snippet network selection to each dataset's characteristics.
For UCF-Crime, which features prominent anomalies, we opted for MGFN due to its proven effectiveness in such scenarios.
In contrast, for VIEW360, which contains more subtle and shorter-duration anomalies, we selected RTFM.
RTFM's approach of training on entire videos minimizes the risk of missing critical frames, making it more suitable for datasets with nuanced anomalies.
For Shanghaitech, which primarily comprises abnormal object appearances, we chose CLIP-TSA to leverage the strong image feature extraction capabilities of CLIP.
This customized strategy enables our FDPN to effectively adapt to diverse anomaly detection scenarios, contributing to its robust performance across different datasets.

\begin{table}[t]
\centering
\small
\vspace{-1em}
\begin{tabular}{@{}p{0.48\linewidth}@{\hspace{0.04\linewidth}}p{0.48\linewidth}@{}}
    \raisebox{-\height}{\adjustbox{max width=\linewidth,valign=t}{%
    \begin{tabular}[t]{c c | c}
    \toprule
    DPS    & Saliency & Direction acc. \\
    \midrule
    \checkmark            &  & 56.50\\
      & \checkmark & 67.80\\
    \checkmark  & \checkmark & \textbf{75.04} \\
    \bottomrule
    \end{tabular}}}
    &
    \raisebox{-\height}{\adjustbox{max width=\linewidth,valign=t}{%
    \renewcommand{\arraystretch}{1.5}%
    \begin{tabular}[t]{l|c|c|c}
    \toprule
    Dataset & SD & OD & Diff. \\
    \midrule
    VIEW360 & 86.00 & 84.78 & -1.24 \\
    UCF-Crime & 88.03 & 86.37 & -1.66 \\
    \bottomrule
    \end{tabular}}} \\[2ex]
    \multicolumn{1}{c}{\centering (a)} & \multicolumn{1}{c}{\centering (b)} \\
\end{tabular}
\vspace{-1em}
\caption{(a) Ablation study for direction prediction on VIEW360,
(b) Effectiveness comparison of image masking process with Saliency Detection (SD) or Object Detection (OD)}
\label{tab:ablation study}
\end{table}

{
\begin{table}[!t]
\centering
\setlength\tabcolsep{6pt}
\begin{adjustbox}{max width=\linewidth} 
\begin{tabular}{c||c|c||c|c||c|c}
\toprule
Grid& \multicolumn{2}{c||}{3x3 (9)} & \multicolumn{2}{|c||}{4x4 (16)} & \multicolumn{2}{|c}{5x5 (25)} \\
\hline
        & Top-K & ROC & Top-K & ROC & Top-K & ROC\\
\cline{2-7}
Unmasked  & 5 / 9 & 85.19          & 10 / 16  & 85.19 & 15 / 25 & 85.02 \\
\cline{2-7}
85.02   & 4 / 9 & \textbf{86.00} & 8 / 16 & 85.50 & 12 / 25 & 85.41 \\
\cline{2-7}
        & 3 / 9 & 85.54          & 6 / 16 & 85.26 & 9 / 25 & 85.30 \\
\bottomrule
\end{tabular}
\end{adjustbox}
\caption{Comparison of different grid sizes and Top-K salient regions for Saliency-driven Image Masking on VIEW360}
\label{tab:Grid_masking_variation}
\vspace{-1em}
\end{table}
}

\begin{figure}[!t]
\centering
\includegraphics[width=\columnwidth]{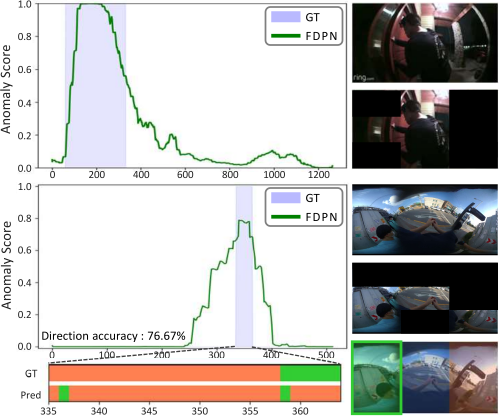}
\caption{
The figure shows FDPN's anomaly scores on the UCF-Crime (top) and VIEW360 datasets (bottom).
Ground truth abnormal frames are highlighted with blue boxes on the graph.
Beside it, we present original and saliency-driven masked images of frames.
In the VIEW360, direction prediction results and ground truth direction are additionally displayed as colored bars:
green for left, blue for center, and orange for right.
   }
\vspace*{-1em}
\label{fig:Qualitative_result}
\end{figure}

\noindent \textbf{Direction Prediction on VIEW360}
Our FDPN also aims to identify the direction of detected abnormal activities to assist visually impaired individuals.
With our DPS, we achieved 75.04\% directional prediction accuracy
by leveraging salient area information from the saliency detector.
We further conduct ablation experiments to understand the impact of each module.
Results in Table~\ref{tab:ablation study}-a demonstrate the essential role of both image features
and salient area information in estimating abnormal event directions,
which highlights their combined effectiveness on VIEW360.

\subsection{Ablation Study}
\noindent \textbf{Saliency Detection vs Object Detection}
As part of our ablation study, we also analyzed the benefits of using a saliency detector
versus a general object detector (YOLO v7 \cite{wang2023yolov7}) in our pre-processing step.
To do this, we replaced the saliency detector as object detector which is trained on COCO dataset.
Results in Table~\ref{tab:ablation study}-b confirmed that the saliency detector is more effective for anomaly detection.

The primary advantage of saliency detection is its focus on active region rather than numerous inactive objects present in video.
Moreover, being class-agnostic, it can detect any active image objects,
enhancing flexibility and robustness.

\noindent \textbf{Optimizing Image Masking}
Our saliency-driven image masking's effectiveness is influenced
by grid size and top-K salient regions retained within the grid, while others are masked.
We thus examined various grid sizes and top-K regions.
Table~\ref{tab:Grid_masking_variation} illustrates this experiment,
emphasizing the critical role of proper masking in highlighting key image areas for more precise detection.
In the table, in a $n \times n$ grid, ``4 / 9'' indicates retaining 4 salient regions out of a total of 9, with 5 areas masked.

\begin{figure}[!t]
\centering
\includegraphics[width=\columnwidth]{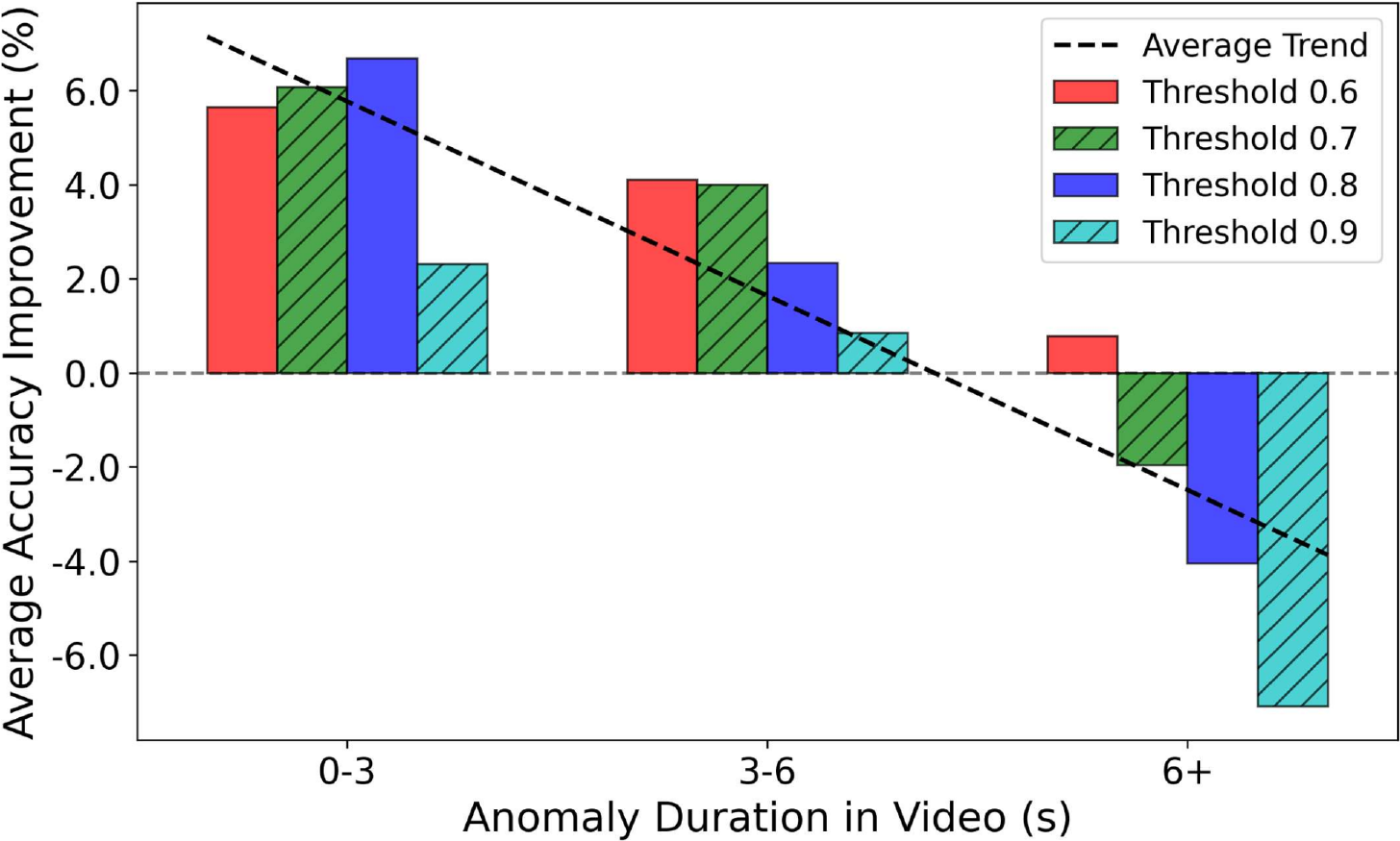}
\caption{
Accuracy improvement of FDPN over MGFN across anomaly durations on UCF-Crime dataset. Thresholds range from 0.6 to 0.9.
Representative abnormal events: 0-3s (RoadAccidents), 3-6s (Abuse, Shooting), 6+s (Assault, Explosion).
   }
\vspace*{-1em}
\label{fig:improvement following duration}
\end{figure}

\noindent \textbf{Verifying Frame-level Prediction's Benefit}
Our FDPN excels in identifying abrupt abnormal events occurring within brief timeframes, thanks to frame-level prediction as illustrated in Figure~\ref{fig:frame_video_comparision}. 
To validate this capability, we conducted a deeper analysis comparing the accuracy improvement of our FDPN against MGFN across varying anomaly durations. 
We evaluated anomaly detection performance using various score thresholds: 0.6, 0.7, 0.8, and 0.9.
The results, presented in Figure~\ref{fig:improvement following duration}, clearly demonstrate that FDPN achieves more substantial improvements for videos containing shorter-duration anomalous events.

\section{Conclusion}
The aim of this research was to tackle safety and security challenges for individuals with visual impairments.
To achieve this, we introduced a novel problem and dataset, \textbf{VIEW360},
aimed at detecting anomalous events and determining their direction
by observing the entire surroundings through an egocentric 360-degree camera.
Additionally, we present \textbf{FDPN}, a new weakly-supervised video anomaly detection method for frame-level detection and direction prediction.
Experimental results show that our FDPN enables more precise detection of anomalous events,
as evidenced by achieving state-of-the-art results on the UCF-Crime, VIEW360 and Shanghaitech datasets.
This method shows promise in enhancing safety for individuals with visual impairments.
We believe our research offers valuable insights for developing AI-based assistive systems,
contributing to their safety and security.

\noindent \textbf{Limitations of our approach}
In consultation with blind individuals and their organizations, our aim was to develop a dataset reflecting real-world situations.
However, it's important to note that our dataset may not cover the full spectrum of challenges faced by the visually impaired,
representing only a selection of potential scenarios.
Additionally, a limitation arises in achieving real-time accessibility,
critical for aiding visually impaired individuals.
Despite our model's accurate anomaly detection capabilities,
its processing speed of 1.7 FPS falls short of real-time requirements.
While other methods operate slightly faster at 2.8 FPS, this still proves inadequate for real-time analysis.
Although this challenge isn't unique to our problem, achieving real-time processing capabilities is crucial for developing assistive technology. Therefore, our future research will focus on improving our model's processing speed while retaining its analytical depth
for providing practical real-time assistance to the visually impaired.
{\small
\bibliographystyle{ieee_fullname}
\bibliography{egbib}
}

\end{document}